\documentclass[conference,letterpaper]{IEEEtran}
\IEEEoverridecommandlockouts
\usepackage[utf8]{inputenc}
\usepackage{cite}
\usepackage{graphicx}
\usepackage{amsmath,amssymb}
\usepackage{url}
\usepackage{booktabs}   
\usepackage{tabularx}   
\usepackage{caption}
\usepackage{siunitx}    
\usepackage[ruled,vlined,linesnumbered]{algorithm2e}
\usepackage{pgfplots}
\pgfplotsset{compat=1.18}
\usepackage{subcaption}

\usepackage{algorithm}
\usepackage{algpseudocode}
\usepackage[dvipsnames]{xcolor}
\usepackage{pgfplots,xcolor}
\pgfplotsset{compat=1.18}

\definecolor{oiBlue}{HTML}{0072B2}
\definecolor{oiOrange}{HTML}{D55E00}

\pgfplotsset{
  groupedbars/.style={
    ybar, bar width=13pt,
    ymin=0,
    enlarge x limits=0.20,
    x=2.2cm,
    xtick=data,
    x tick label style={text width=3.0cm, align=center, font=\footnotesize},
    ytick=\empty, axis line style={draw=none}, tick style={draw=none},
    grid=none,
    clip=true, 
    legend style={
      draw=none, font=\footnotesize, column sep=6pt,
      at={(0.5,0.98)}, anchor=north, legend columns=2 
    },
    nodes near coords,
    nodes near coords align={center},
    point meta=explicit symbolic,
    every node near coord/.style={
      font=\scriptsize\bfseries, text=black, inner sep=0.4pt,
      anchor=south, yshift=0.2ex 
    },
    cycle list={
      {fill=oiBlue,   draw=oiBlue!30!black},
      {fill=oiOrange, draw=oiOrange!30!black},
    }
  }
}

\title{Action Deviation-Aware Inference for Low-Latency Wireless Robots}

\author{%
\IEEEauthorblockN{Jeyoung Park\textsuperscript{1}\textsuperscript{\dag},
Yeonsub Lim\textsuperscript{2}\textsuperscript{\dag},
Seungeun Oh\textsuperscript{2},
Jihong Park\textsuperscript{3},
Jinho Choi\textsuperscript{4},
Seong-Lyun Kim\textsuperscript{2}}
\IEEEauthorblockA{\textsuperscript{1}Department of MME, University of Waterloo, parkjeyoung9@gmail.com}
\IEEEauthorblockA{\textsuperscript{2}School of EEE, Yonsei University, {yeonsub415, seoh, slkim\}@yonsei.ac.kr}
\IEEEauthorblockA{\textsuperscript{3}ISTD Pillar, Singapore University of Technology and Design, jihong\_park@sutd.edu.sg}
\IEEEauthorblockA{\textsuperscript{4}School of EME, University of Adelaide, jinho.choi@adelaide.edu.au}
}
\thanks{
\textsuperscript{\dag}These authors contributed equally to this work.
\newline
\indent This work was supported by Institute of Information \& communications Technology Planning \& Evaluation (IITP) grant funded by the Korea government (MSIT) (No.2022-0-00420, Development of Core Technologies enabling 6G End-to-End On-Time Networking and No.RS-2024-00404972, Development of 5G-A vRAN Research Platform).}
}

\begin{document}


\maketitle

\begin{abstract}
To support latency-sensitive AI applications ranging from autonomous driving to industrial robot manipulation, 6G envisions distributed ML with computational resources in mobile, edge, and cloud connected over hyper-reliable low-latency communication (HRLLC). In this setting, speculative decoding can facilitate collaborative inference of models distributively deployed: a lightweight on-device model locally generates drafts while a more capable remote target model on a server verifies and corrects them in parallel with speculative sampling, thus resulting in lower latency without compromising accuracy. However, unlike autoregressive text generation, behavior cloning policies, typically used for embodied AI applications, cannot parallelize verification and correction for multiple drafts as each generated action depends on observation updated by a previous action. To this end, we propose \textit{Action Deviation-Aware Hybrid Inference (ADAHI)}, wherein drafts are selectively transmitted and verified based on action deviation, which has a strong correlation with action's rejection probability by the target model. By invoking server operation only when necessary, communication and computational overhead can be reduced while accuracy gain from speculative sampling is preserved. Experiments on our testbed show that ADAHI reduces transmission and server operations by approximately 40\%, lowers end-to-end latency by 39.2\%, and attains up to 97.2\% of the task-success rate of baseline that invokes speculative sampling for every draft embedding vector.
\end{abstract}


\section{Introduction}

As Artificial Intelligence (AI) and the Internet of Things (IoT) proliferate, cellular infrastructure will play a critical role in supporting latency-sensitive on-device applications such as cooperative autonomous driving, semantic communication, and robotic manipulation [1]–[4]. However, these applications are not effectively supported by 5G due to its inability to reliably maintain low latency and integrate distributed computing resources [5]. To overcome the practical constraints of 5G in supporting on-device intelligence, the 6G infrastructure envisions enabling latency as low as hundreds of microseconds, data rates of up to 1 Terabit per second, and highly scalable connectivity to support hundreds of thousands of devices simultaneously. By distributing computation across devices and edge servers connected over 6G networks, AI applications will be able to meet latency deadlines that centralized clouds often cannot [6].

With hyper-reliable low-latency communication (HRLLC) that 6G aims to support [7], distributive deployment of AI applications across mobile devices and edge servers will become more prevalent; device-server collaborative inference for large language models (LLMs) and transformers can be facilitated with \textit{speculative decoding} [8] to accelerate inference and reduce communication and computational overhead [9]. Speculative decoding reduces inference latency by having a  more efficient lightweight model---draft model---continuously generate draft tokens, while a more capable larger model---target model---verifies and corrects them in parallel through speculative sampling. This approach is advantageous as the draft and target models can each be deployed to an edge device and a server, respectively, thus efficiently distributing computational resources, and it also prevents excessive communication as the target model only requires the draft model's outputs, unlike split learning necessitating transmission of high-dimensional hidden states [9].

While foundation models, notably LLMs and Vision Language Models (VLMs) with billions of parameters, are anticipated to benefit from the distributed architecture and the 6G's vision of HRLLC [10], collaborative inference can also be adopted for control policies in robotics, used for mapping actions from states and observations. For instance, \textit{behavior cloning} frameworks with transformer architectures such as PerAct [11]  generate probability distributions over \textit{tokens} that represent actions for a given observation, to which speculative sampling can be applied [8]. Behavior cloning, which is a form of \textit{imitation learning} [12], refers to learning robotic skills from expert demonstrations as a supervised learning so that a learned policy can output actions such as joint torques and end-effector poses from observations including camera images and sensor readings [13]. While behavior cloning's use cases range from autonomous driving [1] to robot manipulation [4] and Unmanned Aerial Vehicle (UAV) control [14], policies are almost always required to meet strict end-to-end latency of less than a second [9]. Those for time sensitive platforms such as industrial manipulators and autonomous vehicles often have latency requirements of no more than tens of milliseconds to generate an action [9]. Considering that 6G envisions distributed computing and intelligence and that robotic policies implemented with behavior cloning are inherently on-device, applying collaborative inference with onboard devices and edge servers interconnected by 6G networks would enable meeting both stringent latency and reliability requirements for such applications.

However, while promising, a key challenge to this approach is that speculative decoding cannot directly be applied to the transformers for manipulation tasks. It can straightforwardly accelerate inference for text generation with a target model reviewing the multiple draft tokens in parallel, but this cannot be the case for robotic policies. Unlike language models, which are autoregressive and can reuse their outputs as inputs, action-generation models cannot generate multiple drafts simultaneously because each subsequent action depends on updated observations from executing the previous one. In short, as we look toward the 6G era, we require an architecture that supports speculative decoding in robotic environments.

But what if the on-device draft model can predict whether a certain draft token needs to be verified and corrected by a target model or not? That way, before transmitting, it can predict if the action is likely to be corrected by the target model and transmit only when the draft action requires verification to minimize incurring unnecessary communication and computational overhead. Interestingly, this has already been done in the domain of natural language processing. Uncertainty-Aware Hybrid Language Model (U-HLM) [16]–[17] uses uncertainty---an on-device small language model's (SLM) self-assessed confidence in its outputs with a strong linear relationship with the rejection probability at the server; U-HLM skips uplink transmission for draft tokens with low uncertainty, thus lowering latency from repetitive computation and transmission without compromising the accuracy. 

\begin{figure*}[!t]
  \centering
  \includegraphics[width=\textwidth]{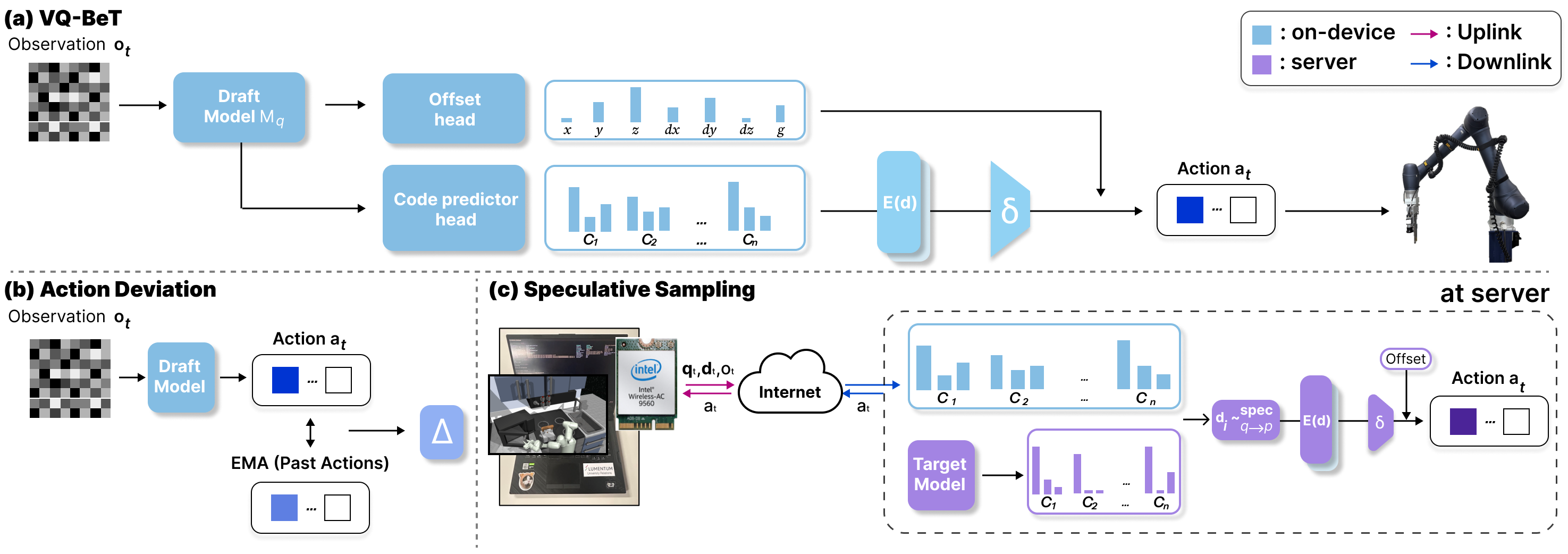}
  \caption{Architecture of Action Deviation-Aware Hybrid Inference. (a) shows VQ-BeT operations: for a given observation $o_t$, the code predictor head generates the probability distribution over the embedding vectors of the codebooks and the offset head, an MLP, computes a small offset [18]. (b) For action generated by on-device VQ-BeT draft model, action deviation $\Delta(t)$ is computed as euclidean distance from exponential moving average of past actions. (c) When $\Delta(t)>\Delta_{th}$, the embedding vectors $d_1 \dots d_n$, the draft model's probability distribution $\mathbf{q}_t$, and observation $o_t$ are transmitted. With these, speculative sampling for each embedding vector occurs and finalized continuous action is transmitted back to the local device.}
  \label{fig:arch}
\end{figure*}

To this end, we present \textit{Action Deviation-Aware Hybrid Inference (ADAHI)}, an architecture to facilitate collaborative inference for action generation with selective speculative sampling. Specifically, given that the draft model must be able to independently evaluate the necessity of speculative sampling at server, we devise \textit{Action Deviation} for estimating the rejection probability for the generated action. Motivated by mean reversion in finance, we compute action deviation by measuring the divergence of the current action from the exponential moving average of the past ones, and we show that it has a strong correlation with the rejection probability at the server. We then apply selective speculative sampling for actions generated by Vector-Quantized Behavior Transformer (VQ-BeT) [18]---a transformer-based architecture for behavior cloning that generates residual embedding vectors for a given observation and reconstructs their sum into a continuous action. As illustrate in Figure~\ref{fig:arch}, a portion of the sampled embedding vectors with high action deviation is transmitted for speculative sampling at server, incurring minimal computational and communication overhead. Through experiments over different use cases ranging from robotic manipulations to swarm control, we demonstrate that our method preserves a target model's performance and effectiveness in action generation, while significantly reducing latency invoked by communication and computation. 


Addressing the absence of framework for collaborative inference in robotics, we envision action generation with distributed computational resources interconnected with low latency wireless networks as a key enabler of embodied on-device intelligence. Our contributions are as follow:
\begin{itemize}
  \item To the best of our knowledge, we are first to apply speculative sampling for robotic motion control, introducing the ADAHI framework that facilitates collaborative inference under latency and accuracy constraints.
  \item We devise and propose \textit{action deviation} for measuring the rejection probability of draft embedding vector indices.
  \item We implement a physical testbed with wireless communication between a local device that deploys a draft model and an edge server hosting a target model.
\end{itemize}

\section{System Model}

VQ-BeT models continuous actions for robotic control by predicting discrete codes in a learned latent action space from observations and optional goal conditions. In this study, we consider a joint utilization of VQ-BeT models varying in sizes, following the architecture in [15]. A locally deployed model serves as a computationally lightweight draft model while its outputs are selectively verified and resampled by a more capable yet compute-intensive target model on server.

\subsection{Vector-Quantized Behavior Transformer}
\label{sec:vqbet}

\textbf{Action Discretization Training.}  During action discretization training, a residual vector-quantized variational autoencoder (RQ-VAE) [19] with an encoder $\varepsilon$ and a decoder $\delta$ learns how to represent actions as latent embedding vectors and how to reconstruct them as continuous actions. First, an action collected from demonstrations is passed through the encoder $\varepsilon$, and the resulting embedding vector is first mapped into the primary codebook $\mathbb{C}_1$ by the nearest neighbor lookup. The residual, which is the remaining error from the quantization, is mapped into subsequent codebooks $\mathbb{C}_2, \ldots, \mathbb{C}_n$, where $\mathbb{C}_\ell = \{{e_1^\ell,...,e_{K}^\ell}\}$, with $e_i^\ell$ denoting the $i$-th embedding vector in codebook $\ell$ and $n$ represents the number of codebooks. $K$ is the number of embedding vectors in the codebook $\ell$, which remains the same across the codebooks. Then, the sum of vectors from the codebooks is passed through the decoder $\delta$ to be reconstructed as the original continuous action. The RQ-VAE is trained with the following loss function: 

\begin{equation}
\begin{split}
L = \| a - \delta(\textstyle\sum_{\ell=1}^n e^\ell_{k}) \|^2 
+ \sum_{\ell=1}^n \| \text{sg}[\epsilon(a)] - e^\ell_{k} \|^2 \\
+ \beta \sum_{\ell=1}^n \| \epsilon(a) - \text{sg}[e^\ell_{k}] \|^2,
\end{split}
\end{equation}

where $a$ is the original action, $\text{sg}[\cdot]$ notates the stop gradient, and $\beta$ is a scalar weight. $\| a - \delta(\textstyle\sum_{\ell=1}^n e^\ell_{k}) \|^2$ is the discrepancy between the decoded action from the sum of the selected embedding vectors and the original action, $\sum_{\ell=1}^n \| \text{sg}[\epsilon(a)] - e^\ell_{k} \|^2$ measures how far the selected code vectors are from the encoder's representation to update the codebook entries, and $\beta \sum_{\ell=1}^n \| \epsilon(a) - \text{sg}[e^\ell_{k}] \|^2$ represents how inconsistent the outputs of the encoder $\varepsilon$ are. 

\textbf{Embedding Vector Prediction.} At step $t$, from a given observation, the transformer via the code predictor head outputs logits $\mathbf{z}_t = \big(z_1(t),\, z_2(t),\, \ldots,\, z_n(t)\big)$ over the learned embedding vectors of codebooks. The logits are normalized into the probability distributions $\mathbf{q}_t = \big(q_1(t),\, q_2(t),\, \ldots,\, q_n(t)\big)$ where the $k$-th element of $q_{\ell}(t)$ is defined as:

\begin{equation}
q^k_{\ell}(t)
= \frac{\exp\!\big(z_\ell^{k}(t)\big)}
       {\sum_{i=1}^{K} \exp\!\big(z_\ell^{k}(t)\big)},
\quad \forall\, k \in \{1,\ldots,K\}.
\end{equation}

The distributions for the codebooks are predicted independently in parallel. Each probability specifies the probability for each index in a codebook to be selected, with these indices denoting discrete latent action primitives---embedding vectors learned by the RQ-VAE, as mentioned earlier. From the distributions $q_1(t),\, q_2(t),\, \ldots,\, q_n(t)$ in $\mathbf{q}(t)$, embedding vectors $d_1(t),\, d_2(t),\, \ldots,\, d_n(t)$ are sampled, and their sum is converted to a continuous action via decoder $\delta$ of the RQ-VAE. 

\textbf{Offset Prediction.} In parallel with the code prediction, the offset head, which is a multilayer perceptron (MLP), predicts a small continuous residual with the same dimensionality as the decoded action from the transformer's hidden state. Once the predicted offset is added to the decoded continuous action, the sampled action is finalized for hardware manipulation.

\textbf{VQ-BeT Training.} The transformer, the code predictor, and offset heads are trained via the loss functions:

\begin{equation}
\begin{split}
L = L_{\text{focal}}\!\left(\zeta_{i=1}^{\text{code}}(o_t)\right) 
+ \beta\, L_{\text{focal}}\!\left(\zeta_{i>1}^{\text{code}}(o_t)\right) \\
+ \big\| a_{t:t+n} - (\lfloor a_{t:t+n} \rfloor + \zeta^{\text{offset}}(o_t)) \big\|_1,
\end{split}
\end{equation}

where $\zeta_{i=1}^{\text{code}}(o_t)$ is the predicted distribution over code indices in $i$ residual codebooks given $o_t$ observation, while $L_{\text{focal}}$ notates the focal loss. Note that the decoder for the reconstruction of the actions from the summed embedding vector remains frozen during the training.


\section{Action Deviation-Aware Hybrid Inference for Vector-Quantized Behavior Transformer}

The primary contribution of this paper is the joint utilization of VQ-BeT models varying in capabilities and locations of deployment using speculative decoding for low latency inference over the wireless networks. Thus, in this section, we demonstrate if the draft model can independently predict the rejection probability of generated embedding vectors with \textit{action deviation} and propose the \textit{ADAHI} framework for selective speculative sampling for actions. 



\subsection{Estimating Rejection Probability Using Action Deviation}

\subsubsection{Action Deviation} For the \textit{t}-th inference, the draft model $M_q$ generates an action $a_t$, and the exponential moving average (EMA) of past actions is computed as follows: 

\begin{equation}
\bar{\mathbf{a}}_t =
\begin{cases}
a_0, & t = 0, \\
(1 - \alpha)\,\bar{\mathbf{a}}_{t-1} + \alpha\,a_{t-1}, & t \ge 1,
\end{cases}
\end{equation}


\noindent where $\alpha$ is the smoothing factor. Action deviation $\Delta(t)$ then is computed as: 

\begin{equation}
\Delta(t) =
\frac{\Delta_{\text{net}}(t)}{\sigma_{\Delta_{net}}}
\end{equation}

\noindent where net deviation is calculated as $\Delta_{\text{net}}(t) = \left\lVert a_t - \bar{\mathbf{a}}_t \right\rVert_2$, while ${\sigma_{\Delta_{net}}}$ is given by $\sqrt{\frac{1}{N} \sum_t \left( \Delta_{\text{net}}(t) - \bar{\Delta}_{\text{net}} \right)^2}$. 

For each use case, we generate more than 50,000 actions. We then partition the range of action deviation into $N$ bins each containing same number of actions and estimate the rejection probability as the fraction of actions with its primary embedding vector probabilistically rejected and resampled by the target model. Only the rejection probability of the primary codebook embedding vector is considered as it defines the fundamental latent representation for structural coherence.

\begin{figure}[!t]
  \centering
  \includegraphics[width=\linewidth]{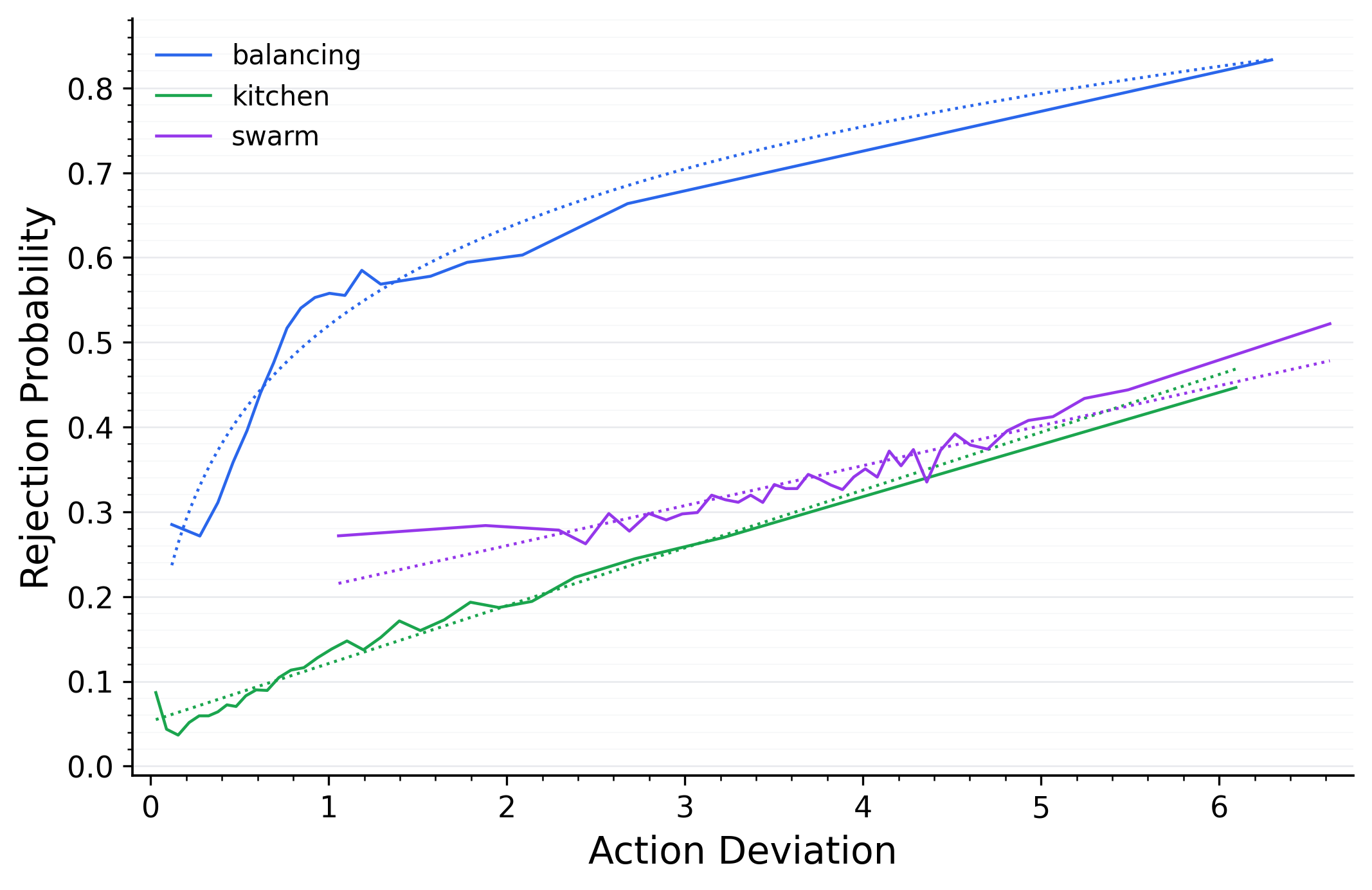}
  \caption{Plots showing the relationship between action deviation and rejection probability for three use cases. Each plot results in correlation coefficient of at least 0.9457 and includes more than 50,000 actions.}
  \label{fig:rejprob}
\end{figure}
\begin{algorithm}[!t]
\caption{Action Deviation-Aware Hybrid Inference}
\label{alg:pda-hi}
\small
\textbf{Input:} Observations $o_t$ \\
\textbf{Output:} executed action $a_t$
\begin{enumerate}
\item $t \leftarrow 0$; 
\item \textbf{repeat}
  \begin{enumerate}
  \item Compute probability distribution $\mathbf{q}_t=M_q(o_t)$. Sample embedding vectors $d_1\dots d_n$ from $\mathbf{q}_t$ and decode draft action $a_t = \delta(\sum_{i=1}^n d_i)$.
  \item Update $\bar a_t=(1-\alpha)\bar a_{t-1}+\alpha a_{t-1}$ and compute $\Delta(t)=\big\|a_t-\bar a_t\|/\sigma_p$.
  \item 
    \begin{enumerate}
    \item \textbf{If} $\Delta \le \Delta_{\mathrm{th}}$ \textbf{then} execute $a_t$.
    \item \textbf{Else}: Transmit $o_t, \mathbf{q}_t, \text{and } d_1\dots d_n$; \\
    Compute probability distribution $\mathbf{p}_t=M_p(o_t)$. \\ 
    For $i=1,\dots,n$: \\
    \text{  }  accept $d_i$ with probability $\min\!\big(1,\tfrac{p_i(d_i)}{q_i(d_i)}\big)$; \\
    \text{  }  if rejected, resample $d_i$ from $\tilde p_i$. \\
    Set $a_t=\delta\!\big(\sum_{i=1}^n d_i\big)+\zeta(o_t)$ and execute $a_t$.
    \end{enumerate}
  \item $t\leftarrow t+1$.
  \end{enumerate}
\item \textbf{until} task complete.
\end{enumerate}
\end{algorithm}
\textbf{Results.} As shown in Figure~\ref{fig:rejprob}, action deviation yields either logarithmic or linear relationship with rejection probability. Based on this, we can build the following model:

\begin{equation}
P_1(t)=
\begin{cases}
m\,\ln\!\bigl(1+\kappa\Delta(t)\bigr)+b \\
m\,\Delta(t)+b 
\end{cases}
\label{eq:rej}
\end{equation}

where $\Delta(t)$ and $P_1(t)$ are action deviation and rejection probability of primary embedding vector at $t$, while $m$, $b$, and $\kappa$ are constants. The ball balancing results in a logarithmic relationship $P_1(t)=0.214\,\ln\!\bigl(1+4.383\Delta(t)\bigr)+0.160$, while the kitchen environment manipulation and the swarm control resulted in linear relationships: $P_1(t)=0.068\,\Delta(t)+0.053$ and $P_1(t)=0.047\,\Delta(t)+0.166$ respectively.

\subsection{Action Deviation-Aware Hybrid Inference}

With the strong correlation shown above, we design the ADAHI. If $\Delta(t) \le \Delta_{th}$, where $\Delta_{\text{th}}$ is a predefined threshold, the action generated by the draft model is accepted for execution. Otherwise, uplink transmission to the server occurs, transmitting the draft's distribution $\mathbf{q}_t = \big(q_1(t),\, q_2(t),\, \ldots,\, q_n(t)\big)$, optional goal conditions, and the embedding vectors indices.  

\subsubsection{Speculative Sampling at Server}\label{subsec:specsamp}

Once the observation, sampled indices, and probability distribution are transmitted the server for  speculative sampling to verify and correct the draft model's output, the target model $M_p$ generates its own probability distributions $\mathbf{p}_t = \big(p_1(t),\, p_2(t),\, \ldots,\, p_n(t)\big)$. Since sampling from each codebook occurs independently, we apply speculative sampling to each codebook's distribution. For the $i$-th codebook, the embedding vector sampled from draft model's distribution is accepted if $q_{i}(t) \le p_{i}$; otherwise, it is rejected with a probability of $1 - \frac{p_{i}(t)}{q_{i}(t)}$, in which case the embedding vector is resampled from an adjusted probability distribution $\mathbf{\tilde{p_i}}(t)$, where the $v$-th probability is given as:


\begin{equation}
\tilde{p_i}_v(t)
= \frac{\max\!\bigl(q_v(t) - p_v(t),\,0\bigr)}
       {\sum_{k=1}^{K}\max\!\bigl(q_k(t) - p_k(t),\,0\bigr)} .
\label{eq:uplink-prob}
\end{equation}

When embedding vector for each codebook is finalized, they are transmitted back to the local device to be summed and constructed into a continuous action by RQ-VAE decoder $\delta$. 

\subsection{Threshold for Skipping Communication and Speculative Sampling}\label{sec:threshold}

\label{sec:experiments}
\begin{table*}[t]
\centering
\vspace*{0.4in}
\caption{Task success rate, mean squared error to ground truth, transmission rate, and true skip ratio of inference methods.}
\setlength{\tabcolsep}{6pt}
\begin{tabular}{@{}lccccccccccccc@{}}
\toprule
& \multicolumn{5}{c}{\textbf{Task Success Rate}} 
& \multicolumn{5}{c}{\textbf{Mean Squared Error}} 
& \multicolumn{3}{c}{\textbf{TR/TSR}}\\
\cmidrule(lr){2-6} 
\cmidrule(lr){7-11}
\cmidrule(lr){12-14}
& \textbf{ADAHI} & \textbf{Hybrid} & \textbf{Draft} & \textbf{Target} & \textbf{Random}
& \textbf{ADAHI} & \textbf{Hybrid} & \textbf{Draft} & \textbf{Target} & \textbf{Random}
& $\mathbf{TR}$ & $\mathbf{TSR}_{\text{Act}}$ & $\mathbf{TSR}_{\text{Rand}}$ \\

\midrule
Kitchen. & \textbf{0.394} & \textbf{0.430} & \textbf{0.060} & \textbf{0.430} & \textbf{0.325} & 1.640 & 1.630 & 2.419 & 1.623 & 1.998 & 0.543 & 0.607 & 0.539 \\
Ball. & \textbf{0.811} & \textbf{0.834} & \textbf{0.001} & \textbf{0.850} & \textbf{0.726} & $5.853{\scriptstyle\text{e-3}}$ & $5.688{\scriptstyle\text{e-3}}$ & $6.116{\scriptstyle\text{e-3}}$ & $5.645{\scriptstyle\text{e-3}}$ & $5.857{\scriptstyle\text{e-3}}$ & 0.603 & 0.553 & 0.343 \\
Swarm. & \textbf{0.294} & \textbf{0.295} & \textbf{0.065} & \textbf{0.303} & \textbf{0.230} & 1.2434 & 1.2359 & 3.9010 & 1.2342 & 2.1291 & 0.591 & 0.504 & 0.395 \\

\bottomrule
\end{tabular}
\label{tab:baseline_perf}
\end{table*}

In this section, we derive the appropriate threshold for skipping, motivated by [15]. To minimize unessential communication and computation for speculative sampling, we can choose to skip for actions with rejection probability lower than a certain threshold. As proven in [8], the outputs sampled from adjusted distribution $\mathbf{\tilde{p}}(t)$ are distributed identically to those from $\mathbf{p}(t)$. With a probabilistic rejection rule, we rearrange the rejection probability of an embedding vector $d_i$ as:

\begin{equation}
\mathbb{E}_{q_{i}(t)}[P_{i}(t)] = \Pr(p_{i}(t)<q_{i}(t))\mathbb{E}_{q_{i}(t)}[1-\frac{p_{i}(t)}{q_{i}(t)} \mid p_{i}(t)<q_{i}(t)]
\end{equation}




\begin{figure}[!t]
  \centering
  \includegraphics[width=\linewidth]{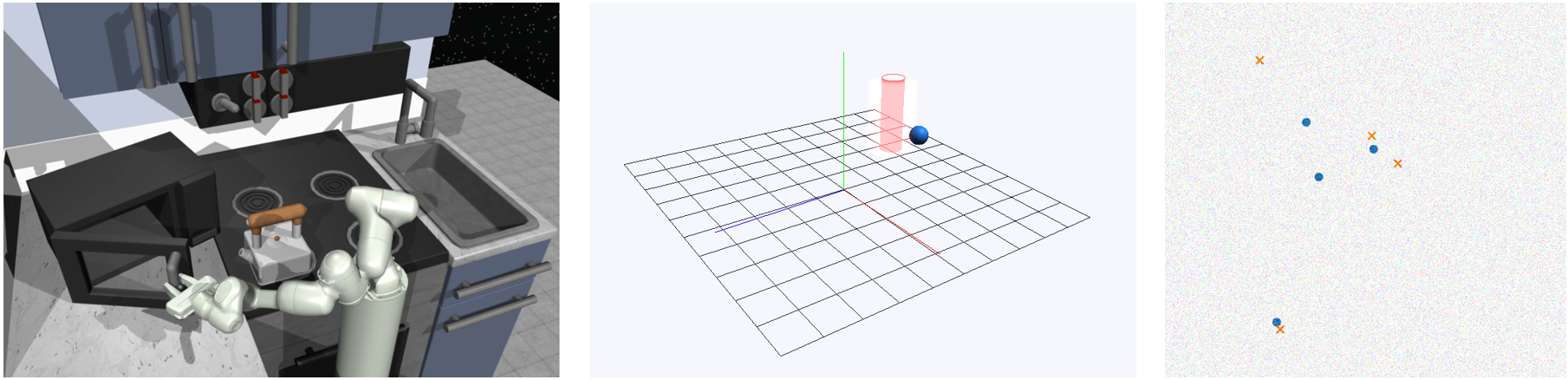}
  \caption{Use cases for ADAHI (from left to right): Kitchen Environment Manipulation, Ball Balancing, and Swarm Control.}
  \label{fig:benchmark}
\end{figure}

Denote by $\tau$ a certain rejection probability threshold. Then our uplink transmission rule is that we transmit if $\tau \le \mathbb{E}_{q_{i}(t)}[P_1(t)]$. Since the action deviation and rejection probability of the primary embedding vector are logarithmically or linearly aligned as shown in Figure~\ref{fig:rejprob}, we approximate the expectation by the value itself as $\mathbb{E}_{q_i(t)}[P_1(t)] \approx P_1(t)$. Using relationships from (\ref{eq:rej}), we can set the threshold to skip speculative sampling for actions with the rejection probability of primary embedding vector less than $\tau$ as:

\begin{equation}
P(t)=
\begin{cases}
\frac{e^{(\tau-b)/m} - 1}{\kappa} \\
\frac{\tau - b}{m}
\end{cases}
\end{equation}

If we were to set a specific transmission rate $TR$, then we can set $TR$ as $\text{Pr}(\Delta > \Delta_{th}) = 1 - F_P(\tau)$, where $F_P$ is the cumulative distribution function of rejection probability. Thus the value of $\tau$ that achieves $TR \in (0,1)$ is $\tau^* = F_P^{-1}(1-TR)$. Then we can use $\tau*$ to facilitate $TR$ as follows:

\begin{equation}
P(t)=
\begin{cases}
\frac{e^{(\tau^*-b)/m} - 1}{\kappa} \\
\frac{\tau^* - b}{m}
\end{cases}
\end{equation}

For the ball balancing, kitchen environment manipulation, and swarm control use cases, we set $\Delta_{th}$ to be 0.890. 0.750, and 3.276, respectively. This choice is based on an ablation study where task success rates remained stable for transmission rate down to 0.6 but declined precipitously afterwards.

\section{Experiments}

With the draft model deployed on the local device and the target model located in the remote server, we conduct experiments to examine the performance of the ADAHI. We  focus on how effectively our method preserves the target model's task success rate and error to ground truth while lowering the latency from communication and computation. 

\textbf{Experimental Setup.} The experiments are conducted on a Windows-based laptop, equipped with a 6-core Intel Core i7-10750H CPU, 8 GB of DDR4 RAM, and an NVIDIA GeForce GTX 1650 Ti GPU. We evaluate action-deviation aware hybrid inference on three simulated use cases, shown in Figure~\ref{fig:benchmark}: (i) Kitchen Environment Manipulation, (ii) Ball Balancing, and (iii) Swarm Control. Kitchen manipulation involves actions including manipulating switch, sliding cabinet, opening microwave, and placing kettle, ball balancing's goal is to tilt the platform to place a ball on a goal location, and swarm control aims to locate multiple drones to target position on 2D space. Baseline methods include draft model only inference, hybrid inference, and random inference. Hybrid inference invokes speculative sampling for every action, while random inference probabilistically transmit actions for speculative sampling, matching the transmission rate of ADAHI.

\textbf{Evaluation Metrics.} The evaluation metrics for measuring the latency and performance of the methods are as follows: Per-Action Latency is computed by dividing the end-to-end latency of the episode by the total number of generated actions; Action throughput represents how many actions does the method generate per second; Transmission Rate (TR) is the probability that an action is transmitted for speculative sampling; True Skip Ratio (TSR) is the probability that an skipped action did not in fact need speculative sampling as it would have been accepted. To measure the performance of the baselines, we compute the task success rate to measure the probability of the goal condition being reached within a given number of steps, and we also measure the mean squared error (MSE) from the ground truth action-observation pairs. 

\textbf{Wireless Communications.} At inference, each draft generated by $M_q$ is paired with its probability distribution, serialized as a JSON object, and sent from the local device to the server deploying the target model $M_p$ via a Flask REST endpoint over the Wi-Fi link. The network switch forwards this JSON payload over Ethernet to the server where the verification and probabilistic resampling of the draft token occur before returning the finalized embedding vectors---again as a JSON object via Flask---over Ethernet and Wi-Fi back to the local device. Experiments use IEEE 802.11ac on the 5 GHz band. The client is a Windows laptop with an Intel Wireless-AC 9560 adapter. During runs, the NIC reported PHY rates around Tx 300 Mb/s and Rx 270 Mb/s with signal 83\% (RSSI -59 dBm). The network performance is average RTT 12.054 ms, RTT jitter 0.302 ms, and downstream throughput 55.4 Mbit/s.

\subsection{Performance and Latency Tradeoffs}\label{sec:perf}

\begin{figure}[!t]
  \centering
  \includegraphics[width=\linewidth]{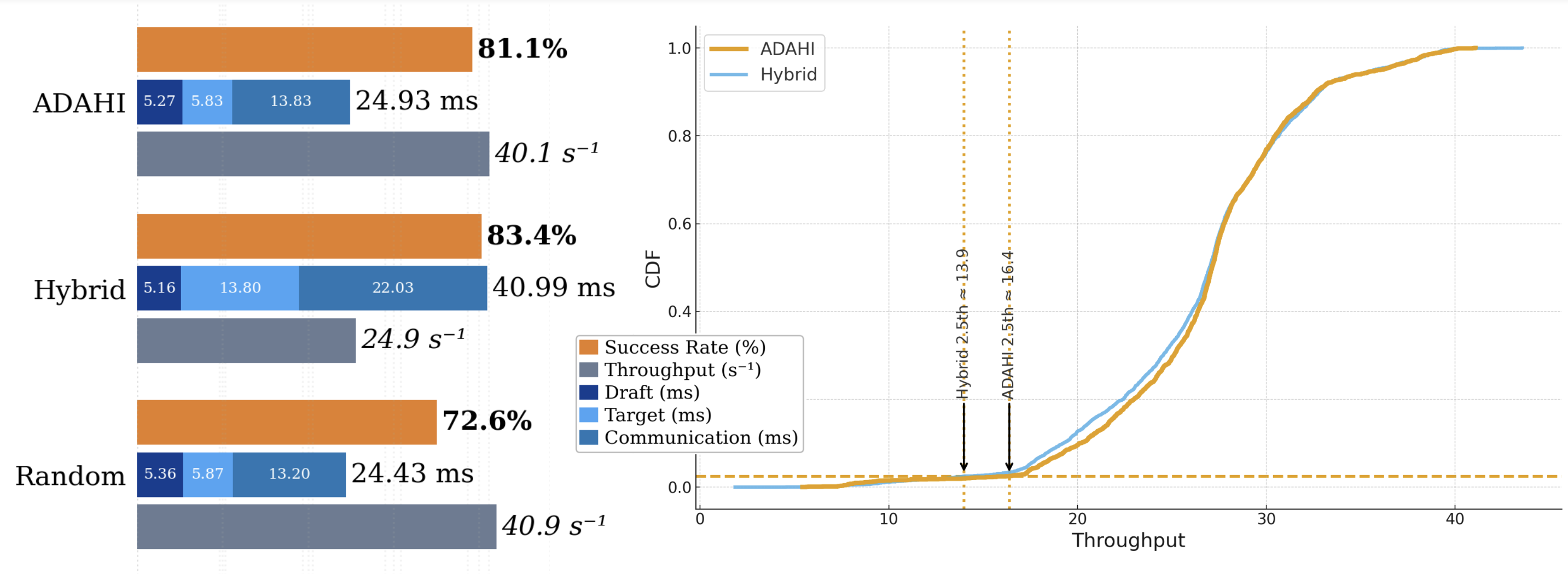}
  \caption{Average per-action latency, task success rate, and action throughput breakdown for inference methods (left); CDF of action throughput for ADAHI and hybrid inference where dotted lines mark each method’s 2.5th percentile (right). }
  \label{fig:perlatency}
\end{figure}

Table~\ref{tab:baseline_perf} and Figure~\ref{fig:perlatency} show the performance and latency tradeoffs of the inference methods. Table~\ref{tab:baseline_perf} demonstrates that ADAHI outperforms random inference across all use cases in terms of task success rate and MSE. In the ball balancing shown in Figure~\ref{fig:perlatency} (left), for example, ADAHI outperforms random inference in terms of task success rate by 10.8\%p, while achieving similar average per-action latency and action throughput. Against hybrid inference, on the other hand, it results in negligible 2.3 \%p drop in task success rate, while significantly lowering latency by 16.06 ms and increasing action throughput by 15.2\text{$s^{-1}$}. Overall, ADAHI, relative to hybrid inference, lowers average per-action latency by 39.2\% and achieved 97.2\% of task success rate of hybrid inference. Figure~\ref{fig:perlatency} (right) indicates a clear advantage in the lower quantiles: at the 2.5th percentile, ADAHI attains higher action throughput of 16.4\text{$s^{-1}$} than 13.9\text{$s^{-1}$} of hybrid inference, evidencing superior worst-case performance. The broad overlap of the CDFs across most quantiles indicates comparable typical performance, while ADAHI’s right shifted onset---its curve begins at a higher throughput---shows it avoids extremely low-throughput cases to result in lower average end-to-end latency. Though less dramatic, in terms of MSE to ground truth action-object pairs, ADAHI similarly locates itself between random inference and target model only inference, while showing only a negligible increase in error compared to target model only inference and hybrid inference. ADAHI's ability to achieve hybrid inference's performance and random inference's latency and throughput can be explained by TR and TSR. As shown in Table~\ref{tab:baseline_perf}, both ADAHI and random inference invoke uplink transmission for speculative sampling for approximately 60\% of the actions; however, ADAHI is more likely to skip ones that are likely to be accepted by the target model, while transmitting those need to be verified more frequently, as it has higher TSR than random inference. Other use cases show a similar trend.

\definecolor{Slate700}{RGB}{15, 53, 87}   
\definecolor{Slate400}{RGB}{107, 124, 147} 
\definecolor{CoolGray550}{RGB}{66, 165, 245} 
\definecolor{Orange}{RGB}{255, 103, 0}

\section{Conclusions}
In this study, we propose Action Deviation-Aware Hybrid Inference to facilitate collaborative inference with speculative decoding for VQ-BeT. By predicting rejection probability with action's Euclidean distance from weighted average of the past, the proposed inference method selectively skips transmission to server and computation for verification and correction, leading to lowering per-action latency while preserving performance. One limitation is that action deviation has only been tested on VQ-Bet. However, considering VQ-BeT is a broadly applicable behavior cloning policy spanning manipulation and navigation, validating on it alone already yields practical value for many applications. While work concentrated on manipulation, we believe this can be extended to other scenarios including autonomous driving and multi-robot deployments.

\end{document}